\documentclass[letterpaper]{article} 
\usepackage[preprint]{aaai2027}  
\usepackage[hyphens]{url}  
\usepackage{graphicx} 
\usepackage{natbib}  
\usepackage{caption} 
\usepackage{amsmath}
\usepackage{amssymb}
\usepackage{algorithm}
\usepackage{algorithmic}
\usepackage{multirow}
\usepackage{booktabs}
\newcommand{\best}[1]{\textbf{#1}}
\newcommand{\second}[1]{\underline{#1}}
\usepackage{newfloat}
\usepackage{listings}
\DeclareCaptionStyle{ruled}{labelfont=normalfont,labelsep=colon,strut=off} 
\floatstyle{ruled}
\newfloat{listing}{tb}{lst}{}
\floatname{listing}{Listing}

\usepackage{booktabs}

\title{FDDWAN: A Frequency-Decoupled Diffusion Network for Watermarking Attack}

\author{
    Chunpeng Wang\textsuperscript{\rm 1},
    Yuxin Li\textsuperscript{\rm 1},
    Xiaoyu Wang\textsuperscript{\rm 2},
    Jidong Yang\textsuperscript{\rm 1},
    Suo Gao\textsuperscript{\rm 3},
    Qi Li\textsuperscript{\rm 1}\corresponding
}

\affiliations{
    \textsuperscript{\rm 1}
    Qilu University of Technology (Shandong Academy of Sciences),
    Jinan, Shandong Province, China\\
    \textsuperscript{\rm 2}
    Dalian Maritime University,
    Dalian, Liaoning Province, China\\
    \textsuperscript{\rm 3}
    Dalian Polytechnic University,
    Dalian, Liaoning Province, China\\[3pt]
}

\begin{document}

\maketitle

\begin{abstract}
 Existing invisible watermark removal methods often struggle to accurately capture the watermark-bearing features, leading to an unfavorable trade-off between watermark suppression and perceptual fidelity. In this paper, we propose the Frequency-Decoupled Diffusion Watermark Attack Network (FDDWAN), a coarse-to-fine framework that performs watermark removal through wavelet-domain decomposition and residual diffusion refinement. In the initial stage, the Wavelet-based Frequency-domain Preliminary Attack Module (WFPAM) decomposes the watermarked image into low- and high-frequency subbands and applies frequency-specific attack strategies tailored to their respective contributions to watermark robustness and perceptual quality. In the next stage, the Frequency-domain Residual Diffusion Attack Module (FRDAM) separately models the residual distributions between the preliminarily attacked outputs and the corresponding watermark-free references during training. Rather than reconstructing the entire image, FRDAM selectively refines frequency-domain residuals, directing the diffusion process toward the remaining watermark-related discrepancies while minimizing modifications to image content. Extensive experiments on CelebA and ImageNet across four representative watermarking schemes demonstrate that FDDWAN achieves a more favorable trade-off between watermark removal effectiveness and visual fidelity than conventional and learning-based attack methods.
\end{abstract}
\begin{figure*}[t]
    \centering
    \includegraphics[width=1.0\textwidth]{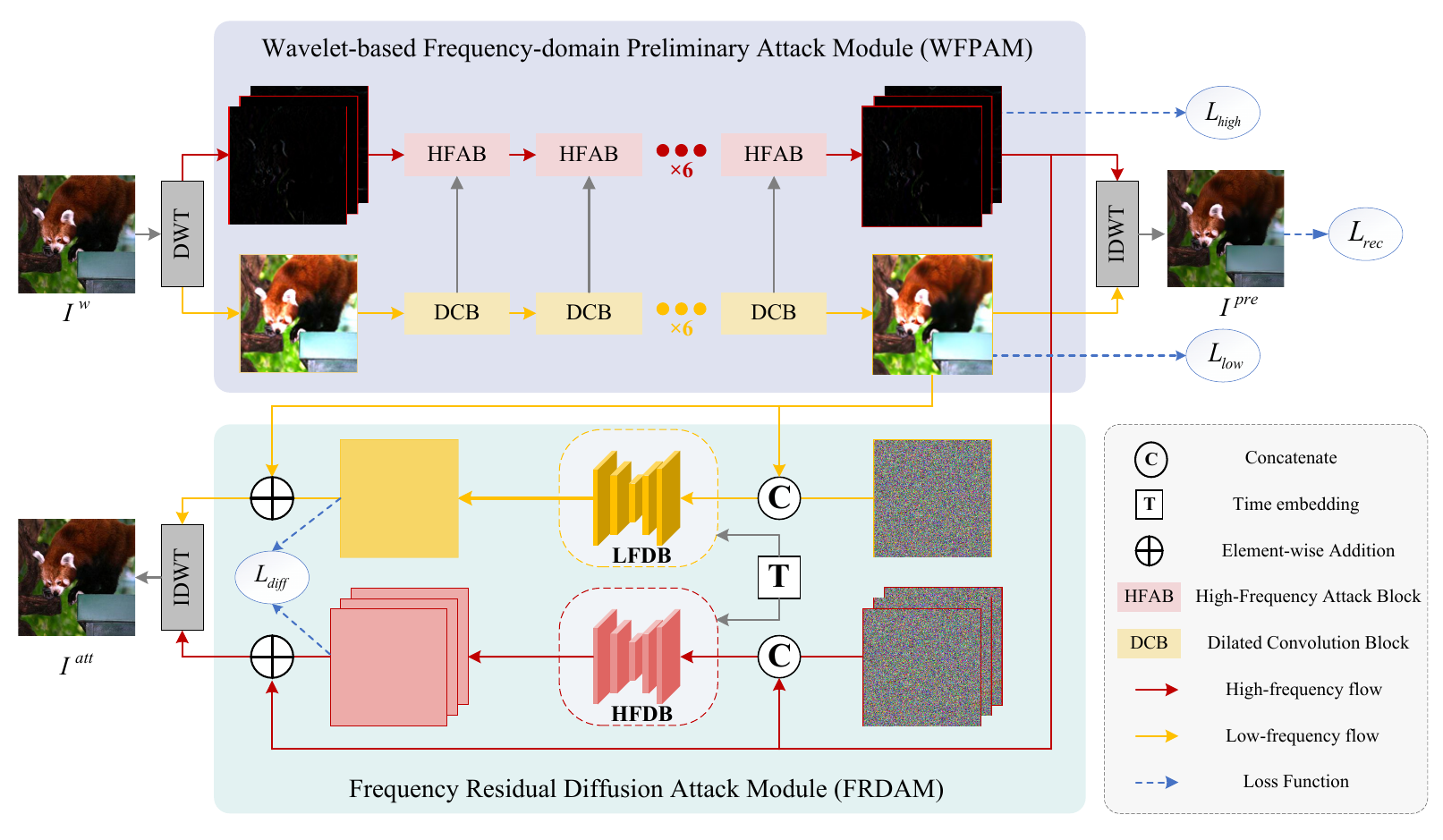}
    \caption{Overall framework of FDDWAN. It contains two detachable modules: Wavelet-based Frequency-domain Preliminary Attack Module (WFPAM) and Frequency Residual Diffusion Attack Module (FRDAM).}
    \label{fig:framework}
\end{figure*}
\section{Introduction}
Invisible image watermarking embeds imperceptible metadata into visual content and has served as a foundational technology for various security tasks, including copyright protection~\cite{kou2025iwrn}, ownership verification~\cite{wang2024must}, and forgery detection~\cite{liu2023fragile}. The robustness of a watermarking system, however, ultimately depends on whether the watermark remains reliably recoverable after routine post-processing or adversarial manipulation~\cite{Chen_2026_CVPR}. This motivates the study of invisible watermark removal, where an attacker seeks to disrupt or invalidate watermark recovery while preserving the perceptual fidelity of the watermarked image. Therefore, investigating such attack methods is essential for clarifying the practical security boundaries of existing watermarking schemes, revealing their underlying vulnerabilities, and informing the design of more robust watermarking defenses.

Existing watermark removal methods still struggle with the trade-off between attack ability and perceptual fidelity. Conventional attacks, such as noise addition, filtering, compression, and geometric transformation, are simple but input-agnostic, often causing substantial image degradation before invisible watermarks can be effectively destroyed. Recent learning-based watermark removal methods have effectively alleviated the limitation of conventional attacks. Nevertheless, many of them operate directly in the spatial domain, where watermark-related signals are intricately entangled with semantic structures and visual details. Consequently, these methods may either fail to eliminate recoverable watermark evidence or introduce perceptible artifacts into the attacked image. A key insight is that invisible watermark exhibit distinct frequency-domain characteristics: low-frequency features are usually exploited to enhance robustness, whereas high-frequency features predominantly encode textures, edges, and other fine details for reconstruction. This frequency dependence indicates that effective watermark removal should selectively manipulate frequency features rather than apply uniform perturbations in the spatial domain.

In this paper, we propose a Frequency-Decoupled Diffusion Watermarking Attack Network (FDDWAN) for invisible watermark removal, as shown in Fig.~\ref{fig:framework}. FDDWAN consists of two sequential modules: a Wavelet-based Frequency-domain Preliminary Attack Module (WFPAM) and a Frequency-domain Residual Diffusion Attack Module (FRDAM). WFPAM performs a targeted preliminary attack by separately processing the low- and high-frequency wavelet subbands, while FRDAM further removes the remaining watermark traces through diffusion-based residual refinement in the wavelet domain. By combining frequency-specific perturbation with
residual diffusion modeling, FDDWAN achieves a favorable trade-off between watermark removal effectiveness and image fidelity. Our contributions are
summarized as follows:

\setlength{\parindent}{1em}(1) We formulate invisible watermark removal as a frequency-decoupled attack problem and propose a two-stage framework that integrates wavelet-domain initial suppression with diffusion-based residual refinement.

\setlength{\parindent}{1em}(2) We introduce a frequency-domain residual diffusion strategy that separately models the residual distributions of low- and high-frequency subbands, enabling more effective watermark suppression while preserving semantic structures and texture details for reconstruction.

\setlength{\parindent}{1em}(3) Extensive experiments on CelebA and ImageNet under multiple watermarking schemes demonstrate that FDDWAN achieves a better balance between watermark removal effectiveness and visual quality, outperforming representative conventional and learning-based attack methods.
\section{Related Work}

\subsection{Invisible watermarking}

Invisible image watermarking aims to embed imperceptible ownership or authentication information into visual content while maintaining perceptual quality and watermark robustness. Early watermarking methods primarily relied on handcrafted embedding strategies and can generally be categorized into spatial-domain, transform-domain, and moment-based approaches. Spatial-domain methods, such as LSB, SIRD, and SDC~\cite{chan2004lsb,abraham2019imperceptible,luo2010sdc}, encode watermark information by directly modifying pixel values. Transform-domain schemes instead embed watermarks into coefficients derived from the DFT, DCT, DWT-DCT, or related transforms~\cite{urvoy2014dft,sisaudia2022dct,alhaj2007dwtdct}, generally providing greater robustness against common signal processing attacks. Moment-based approaches, including PHFMs, RHFMs, and PHTs ~\cite{phfm,RHFM,PHTs}, exploit invariant representations to further improve the resistance to rotation, scaling, and other geometric distortions.

More recently, deep learning-based watermarking methods have used neural networks to jointly optimize watermark embedding and extraction. MBRS~\cite{jia2021mbrs} improves robustness against JPEG compression by incorporating both real and simulated compression data during training. HiDDeN~\cite{zhu2018hidden} formulates data hiding and blind watermarking within an end-to-end learning framework, while StegaStamp~\cite{tancik2020stegastamp} uses an encoder--decoder network to embed imperceptible bit strings and improves decoding robustness by simulating printing, photography, and geometric distortions during training. TrustMark~\cite{bui2025trustmark} combines spatial and spectral objectives to achieve robust watermarking for images of arbitrary resolutions. These methods generally follow a post-hoc setting~\cite{an2024waves}, in which watermarks are embedded into existing images. In-generation watermarking instead incorporates watermark information directly into the image synthesis process. Stable Signature~\cite{fernandez2023stable} fine-tunes the decoder of a latent diffusion model so that generated images contain a predefined binary signature for source attribution. Compared with conventional approaches, learning-based watermarking methods can learn more adaptive embedding strategies and achieve better robustness across diverse distortions.
\subsection{Invisible watermarking Attack}
Invisible watermark removal aims to disrupt embedded watermark information while preserving the perceptual quality of the attacked image. Conventional attacks rely on signal processing operations, such as noise addition, filtering, compression, cropping, and geometric transformations~\cite{wang2024cropping}. However, these non-selective perturbations often provide limited effectiveness against robust watermarks or introduce noticeable visual degradation~\cite{wang2007att1,DAOUI2022att2}.

Learning-based methods improve attack adaptivity by learning transformations from watermarked images to attacked outputs. FAADW~\cite{chen2025faadw} employs feature attention to extract watermark-related residual features, whereas HIWANet~\cite{wang2024hiwanet} combines asymmetric optimization with high-frequency suppression to reduce perceptual distortion. Diffusion-based methods have also been studied for watermark removal. DiffWA~\cite{li2023diffwa} treats watermark removal as a conditional reconstruction task. Zhao et al.~\cite{zhao2024vaeattack} propose a family of regeneration attacks that add Gaussian noise to watermarked images or their latent representations and then reconstruct them using pretrained generative models. The two variants, VAEAttack and Diffusion Attack, use pretrained VAEs and diffusion models, respectively. In addition, UnMarker~\cite{kassis2025unmarker} disrupts robust watermarks by directly optimizing the image spectrum without requiring access to the watermark detector. Despite their effectiveness, existing methods generally process the image in the spatial domain, regenerate the image as a whole, or manipulate frequency information without explicitly modeling the distinct roles of different frequency bands. In contrast, FDDWAN performs frequency-specific preliminary attacks and residual diffusion refinement in the wavelet domain, thereby reducing unnecessary modifications to image content.
\section*{Method}

Given a watermarked image $I^{w}\in\mathbb{R}^{H\times W\times c}$, our goal is to generate an attacked image $I^{att}\in\mathbb{R}^{H\times W\times c}$ that disrupts watermark extraction while maintaining perceptual fidelity and semantic consistency with the input image. FDDWAN explicitly decouples low- and high-frequency subbands, performs a targeted preliminary attack in the wavelet domain, and then removes watermark information through diffusion-based residual refinement. As shown in Fig.~\ref{fig:framework}, FDDWAN contains two sequential modules. The Wavelet-based Frequency-domain Preliminary Attack Module (WFPAM) performs the preliminary attack by perturbing low- and high-frequency subbands with different strategies. The Frequency-domain Residual Diffusion Attack Module (FRDAM) then models the residual between the preliminarily attacked output and the original image in the wavelet domain.
\subsection*{Wavelet-Domain Representation}
Wavelet transforms decompose images into localized frequency subbands
while preserving spatial characteristics and multi-scale information~\cite{xiang2025wavelet}.
Invisible watermarks are typically distributed across frequency bands with different functional roles: low-frequency subbands contribute primarily to robustness against common distortions, whereas high-frequency features facilitate imperceptibility by embedding watermarks within textures and structural details. Motivated by these observations, we apply the discrete wavelet transform (DWT) to decompose the watermarked image $I^{w}$.
\begin{equation}
\operatorname{DWT}(I^{w})=(I_{LL}^{w},I_{LH}^{w},I_{HL}^{w},I_{HH}^{w}),
\end{equation}
where $I_{LL}^{w}$ denotes the low-frequency approximation subband,
while $I_{LH}^{w}$, $I_{HL}^{w}$, and $I_{HH}^{w}$ denote the
horizontal, vertical, and diagonal detail subbands, respectively.

After the attack, the image will be  reconstructed by inverse DWT(IDWT):
\begin{equation}
I^{att}=\operatorname{IDWT}(I_{i}^{pre}+\hat{r}_{i}),
\end{equation}
where $I_{i}^{pre}$ denotes the preliminarily attacked subbands generated by WFPAM and $\hat{r}_{i}$ denotes the frequency residuals predicted by FRDAM, where $i\in\{LL,LH,HL,HH\}$.
\begin{figure}[t]
\centering
\includegraphics[width=1.0\columnwidth]{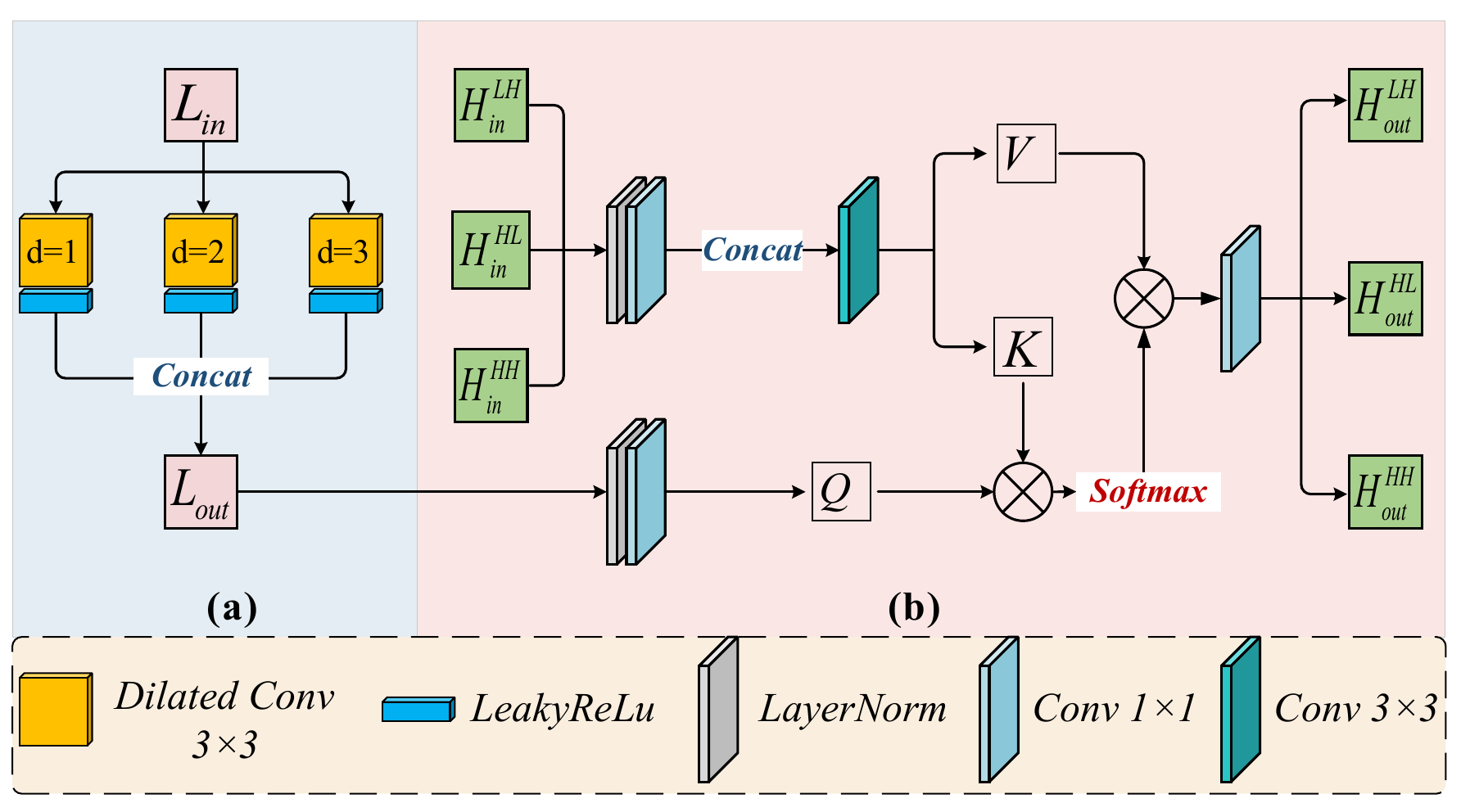}
\caption{Architecture of (a) the Dilated Convolution Block and (b) the High-Frequency Attack Block.}
\label{fig:fig2}
\end{figure}
\subsection{Wavelet-based Frequency Preliminary Attack Module}
WFPAM aims to generate a preliminarily attacked image in the wavelet domain. WFPAM comprises two coordinated branches which consists of six Dilated Convolution Blocks (DCBs) and six High-Frequency Attack Blocks (HFABs) respectively. The DCBs process the low-frequency subband, focusing on destroying invisible watermark information embedded into the low-frequency regions, while the HFABs suppresses high-frequency watermark cues under the guidance of low-frequency structural feature. The six DCBs are paired with six HFABs. Within each pair, the DCB output is fed into the HFAB as low-frequency structural guidance.
\subsubsection{Dilated Convolution Block.}To capture watermark-related features at different spatial scales, each DCB uses three parallel dilated convolutions, as shown in Fig.~\ref{fig:fig2}(a). Given an input low-frequency feature $L_{in}$, the output of the $i$-th branch is
\begin{equation}
L^{(i)}=\operatorname{LeakyReLU}(L_{in}*_{d_i}W^{(i)}),i=1,2,3
\end{equation}
where $d_i\in\{1,2,3\}$ is the dilation rate, $*_{d_i}$ represents a 3x3 convolution and $W^{(i)}$ denotes the learnable convolution kernel. The outputs of the three branches are then concatenated along the channel dimension
\begin{equation}
L_{out}=\operatorname{Concat}(L^{(1)},L^{(2)},L^{(3)}).
\end{equation}

The feature $L_{out}$ is passed to the next DCB and simultaneously used as the query corresponding HFAB. After six DCBs, we denote the preliminarily attacked low-frequency subband by $I_{LL}^{pre}$.
\subsubsection{High-Frequency Attack Block.}As shown in Fig.~\ref{fig:fig2}(b). Given three input high-frequency features $H_{in}^{LH}$, $H_{in}^{HL}$, and $H_{in}^{HH}$, the HFAB first concatenates them along the channel dimension: $H_{in}=\operatorname{Concat}(H_{in}^{LH},H_{in}^{HL},H_{in}^{HH})$. 
The query is generated from $L_{out}$, while the key and value are generated from $H_{in}$:

\begin{equation}
\begin{aligned}
Q&=W_Q\cdot \operatorname{LN}(L_{out}),\\
K&=W_K\cdot \operatorname{Conv}_{3\times3}\bigl(\operatorname{Conv}_{1\times1}(\operatorname{LN}(H_{in}))\bigr),\\
V&=W_V\cdot \operatorname{Conv}_{3\times3}\bigl(\operatorname{Conv}_{1\times1}(\operatorname{LN}(H_{in}))\bigr),
\end{aligned}
\end{equation}
where $\operatorname{LN}(\cdot)$ denotes layer normalization, $W_Q$, $W_K$, and $W_V$ are learnable projections used to generate the query $Q$, key $K$, and value $V$, respectively.

Then, the output feature map can be obtained from the formula:

\begin{equation}
A=\operatorname{Softmax}\left(\frac{QK^{T}}{\sqrt{d_k}}\right)V,
\end{equation}
where $d_k$ denotes the feature dimension of the projected query and key vectors.

Finally, the attention feature $A$ is projected by a $1\times1$ convolution and then divided along the channel dimension to obtain the three directional high-frequency outputs:

\begin{equation}
(H_{out}^{LH},H_{out}^{HL},H_{out}^{HH})=\operatorname{split}(\operatorname{Conv}_{1\times1}(A)),
\end{equation}
where $\operatorname{Conv}_{1\times1}(\cdot)$ maps the attention feature $A$ to $3c$ channels, and $\operatorname{split}(\cdot)$ divides the result into three $c$-channel features corresponding to the $LH$, $HL$, and $HH$ directions.

After six paired DCBs and HFABs, we obtained the preliminarily attackedsubbands $I_{LL}^{pre}$, $I_{LH}^{pre}$, $I_{HL}^{pre}$, and $I_{HH}^{pre}$. Through IDWT, we can obtain the preliminarily attackedimage:

\begin{equation}
I^{pre}=\operatorname{IDWT}(I_{LL}^{pre},I_{LH}^{pre},I_{HL}^{pre},I_{HH}^{pre}).
\end{equation}
\subsection{Frequency-Domain Residual Diffusion Attack Module}
FRDAM learns the residual corrections between preliminarily attacked subbands and their corresponding clean references, allowing the diffusion model to focus on frequency-domain refinement rather than full-subbands generation.
FRDAM consists of two separately parameterized conditional diffusion
branches: the Low-Frequency Diffusion Branch (LFDB) and the
High-Frequency Diffusion Branch (HFDB). LFDB processes the $LL$
subband, whereas HFDB uses one shared parameter set to process the
$LH$, $HL$, and $HH$ subbands. For clarity, the diffusion process is
described below using LFDB as an example.

The target low-frequency residual and its conditional subbands are
defined as:
\begin{equation}
x_0^{(l)}
=
r_{LL}
=
G_{LL}-I_{LL}^{\mathrm{pre}},
\qquad
x_c^{(l)}
=
I_{LL}^{\mathrm{pre}},
\label{eq:frdam_target}
\end{equation}
where $G_{LL}$ and $I_{LL}^{\mathrm{pre}}$ denote the clean and
preliminarily attacked low-frequency subbands, respectively.
$x_0^{(l)}$ is the target residual modeled by LFDB, while
$x_c^{(l)}$ is the conditional input supplied to the denoising network.
The clean subband $G_{LL}$ is used only during training to construct
the target residual.

\subsubsection{Forward Process.} 
Following DDPM~\cite{ho2020ddpm}, the forward process gradually adds
Gaussian noise to the target residual. Its one-step transition is
defined as:
\begin{equation}
q\left(
x_t^{(l)}
\mid
x_{t-1}^{(l)}
\right)
=
\mathcal{N}\left(
x_t^{(l)};
\sqrt{\alpha_t}\,x_{t-1}^{(l)},
(1-\alpha_t)\mathbf{I}
\right),
\label{eq:frdam_forward_step}
\end{equation}
where $\alpha_t=1-\beta_t$, $\beta_t$ controls the noise variance at
diffusion timestep $t$, and $t\in\{1,\ldots,T\}$, with $T$ denoting
the total number of diffusion timesteps.

Let $\bar{\alpha}_t=\prod_{\tau=1}^{t}\alpha_{\tau}$. The noisy residual
at an arbitrary step $t$ can be obtained directly from $x_0^{(l)}$ as:
\begin{equation}
q\left(
x_t^{(l)}
\mid
x_0^{(l)}
\right)
=
\mathcal{N}\left(
x_t^{(l)};
\sqrt{\bar{\alpha}_t}\,x_0^{(l)},
(1-\bar{\alpha}_t)\mathbf{I}
\right).
\label{eq:frdam_forward_closed}
\end{equation}
Equivalently, the noisy residual is sampled by:
\begin{equation}
x_t^{(l)}
=
\sqrt{\bar{\alpha}_t}\,x_0^{(l)}
+
\sqrt{1-\bar{\alpha}_t}\,\epsilon^{(l)},
\qquad
\epsilon^{(l)}
\sim
\mathcal{N}(\mathbf{0},\mathbf{I}).
\label{eq:frdam_forward_sample}
\end{equation}

\subsubsection{Reverse Process.} The reverse process starts from
$x_T^{(l)}\sim\mathcal{N}(\mathbf{0},\mathbf{I})$ and progressively
recovers the target residual under the guidance of $x_c^{(l)}$.
At each reverse step, the conditional U-Net takes the noisy residual
$x_t^{(l)}$, the conditional component $x_c^{(l)}$, and the timestep
$t$ as inputs, and predicts the Gaussian noise contained in
$x_t^{(l)}$:
\begin{equation}
\hat{\epsilon}_t^{(l)}
=
\epsilon_{\theta_L}
\left(
x_t^{(l)},x_c^{(l)},t
\right),
\label{eq:frdam_noise_prediction}
\end{equation}
where $\theta_L$ denotes the parameters of LFDB. The network directly
predicts only $\hat{\epsilon}_t^{(l)}$. Based on this prediction, the
target residual at the current reverse step is analytically estimated
as:
\begin{equation}
\hat{x}_{0,t}^{(l)}
=
\frac{
x_t^{(l)}
-
\sqrt{1-\bar{\alpha}_t}\,
\hat{\epsilon}_t^{(l)}
}{
\sqrt{\bar{\alpha}_t}
}.
\label{eq:frdam_residual_estimation}
\end{equation}

During inference, we employ deterministic DDIM sampling~\cite{ddim}
with $\eta=0$. Let
$\mathcal{T}=\{t_1,t_2,\ldots,t_K\}$ denote a decreasing sequence of
selected sampling timesteps, where
$t_1>t_2>\cdots>t_K$, and let $\bar{\alpha}_{t_{K+1}}=1$.
The deterministic DDIM update from $t_k$ to $t_{k+1}$ is
\begin{equation}
x_{t_{k+1}}^{(l)}
=
\sqrt{\bar{\alpha}_{t_{k+1}}}\,
\hat{x}_{0,t_k}^{(l)}
+
\sqrt{1-\bar{\alpha}_{t_{k+1}}}\,
\hat{\epsilon}_{t_k}^{(l)}.
\label{eq:frdam_ddim_update}
\end{equation}

In our implementation, each diffusion branch employs $K=10$ selected
sampling timesteps. Starting from $x_{t_1}^{(l)}$, 
Eq.~\eqref{eq:frdam_ddim_update} is iteratively applied along the
selected timestep sequence. At the end, the terminal estimate
$\hat{x}_0^{(l)}$ is the low-frequency residual predicted by LFDB:
\begin{equation}
\hat{r}_{LL}
=
\hat{x}_0^{(l)}.
\label{eq:frdam_lf_residual}
\end{equation}

HFDB follows the same forward and reverse processes for the $LH$, $HL$,
and $HH$ subbands. The three high-frequency subbands share a single
parameter set $\theta_H$, and the terminal outputs of their reverse
processes are used as the corresponding predicted residuals:
\begin{equation}
\hat{r}_{j}=\hat{x}_0^{(j)},
\qquad
j\in\{LH,HL,HH\}.
\label{eq:frdam_hf_residuals}
\end{equation}
\subsection*{Loss Functions}
\subsubsection{Preliminary Attack Loss.}
Let $G$ denote the clean host image, and let
$G_{LL}$, $G_{LH}$, $G_{HL}$, and $G_{HH}$ denote its wavelet
subbands. WFPAM is supervised directly in the wavelet domain by aligning
the preliminarily attacked subbands with the corresponding clean subbands.
The low-frequency loss is defined as:
\begin{equation}
\mathcal{L}_{\mathrm{low}}
=
\left\|
I_{LL}^{pre}-G_{LL}
\right\|_{2},
\label{eq:low_loss}
\end{equation}

The high-frequency loss is computed over the three directional
high-frequency subbands:
\begin{equation}
\mathcal{L}_{\mathrm{high}}
=
\sum_{j\in\{LH,HL,HH\}}
\left\|
I_{j}^{pre}-G_{j}
\right\|_{2},
\label{eq:high_loss}
\end{equation}
where $I_{j}^{pre}$ and $G_j$ denote the preliminarily attacked and
clean high-frequency subbands in direction $j$, respectively.

To further preserve image content in the spatial domain, we employ
a mean loss in L1 reconstruction:
\begin{equation}
\mathcal{L}_{\mathrm{rec}}
=
\frac{1}{N}
\left\|
I^{\mathrm{pre}}-G
\right\|_{1},
\label{eq:rec_loss}
\end{equation}
where $I^{\mathrm{pre}}$ is the preliminarily attacked image reconstructed
from the four preliminarily attacked subbands, and $N$ is the total number
of image elements.

The overall preliminary attack loss is defined as:
\begin{equation}
\mathcal{L}_{\mathrm{pre}}
=
\lambda_{\mathrm{low}}\mathcal{L}_{\mathrm{low}}
+
\lambda_{\mathrm{high}}\mathcal{L}_{\mathrm{high}}
+
\lambda_{\mathrm{rec}}\mathcal{L}_{\mathrm{rec}},
\label{eq:pre_loss}
\end{equation}
where $\lambda_{\mathrm{low}}$, $\lambda_{\mathrm{high}}$, and
$\lambda_{\mathrm{rec}}$ control the contributions of the
low-frequency loss, high-frequency loss, and spatial reconstruction
loss, respectively.
\subsubsection{Diffusion Loss.}
FRDAM is trained using the noise-prediction objective. The three
high-frequency subbands are stacked along the batch dimension and
processed by HFDB with shared parameters $\theta_H$. The overall
diffusion loss is defined as:
\begin{equation}
\begin{aligned}
\mathcal{L}_{\mathrm{diff}}
={}&
\mathbb{E}_{t,\epsilon^{(l)}}
\left[
\left\|
\epsilon^{(l)}
-
\epsilon_{\theta_L}
\left(
x_t^{(l)},x_c^{(l)},t
\right)
\right\|_2^2
\right]
\\
&+
\mathbb{E}_{t,\epsilon^{(h)}}
\left[
\left\|
\epsilon^{(h)}
-
\epsilon_{\theta_H}
\left(
x_t^{(h)},x_c^{(h)},t
\right)
\right\|_2^2
\right],
\end{aligned}
\label{eq:diff_loss}
\end{equation}
where the superscripts $(l)$ and $(h)$ denote the low-frequency sample
and the batch-stacked high-frequency samples, respectively.

\subsubsection{Total Loss.} WFPAM and FRDAM are jointly optimized using the following overall
training objective:
\begin{equation}
\mathcal{L}_{\mathrm{total}}
=
\mathcal{L}_{\mathrm{pre}}
+
\lambda_{\mathrm{diff}}\mathcal{L}_{\mathrm{diff}},
\label{eq:total_loss}
\end{equation}
where $\lambda_{\mathrm{diff}}$ controls the contribution of the
diffusion loss. During joint training,
$\mathcal{L}_{\mathrm{pre}}$ supervises the preliminary frequency-domain
attack and spatial reconstruction, while $\mathcal{L}_{\mathrm{diff}}$
trains LFDB and HFDB to predict the noise added to the corresponding
low- and high-frequency residual targets.
\section{Experimental Results}
\begin{table*}[t]
\centering
\small
\setlength{\tabcolsep}{0.3pt}
\renewcommand{\arraystretch}{1.3}

\begin{tabular*}{\textwidth}{
  @{}cc
  @{\extracolsep{\fill}}
  *{4}{c}
  @{\hspace{5pt}}
  *{4}{c}
  @{}
}
\hline
\multicolumn{2}{c}{\multirow{2}{*}{Attack method}}
& \multicolumn{8}{c}{Watermarking method} \\
\cline{3-10}

\multicolumn{2}{c}{}
& \multicolumn{4}{c}{CelebA dataset (PSNR/BER)}
& \multicolumn{4}{c}{ImageNet dataset (PSNR/BER)} \\
\cline{3-6}\cline{7-10}

\multicolumn{2}{c}{}
& DCT & PHFMs & HiDDeN & StegaStamp
& DCT & PHFMs & HiDDeN & StegaStamp \\
\hline

\multirow{5}{*}{\rotatebox[origin=c]{90}{Conventional}}
& Gaussian noise
& 30.09/0.1182
& 27.02/0.0352
& 30.04/0.0644
& 32.52/0.0333
& 30.20/0.1025
& 30.21/0.0195
& 30.18/0.0574
& 30.13/0.0281 \\

& Salt \& pepper
& 27.65/0.1553
& 27.36/0.0742
& 28.13/0.0458
& 27.29/0.0018
& 27.36/0.0537
& 27.44/0.0273
& 27.93/0.0788
& 28.25/0.0021 \\

& Speckle noise
& 29.41/0.1260
& 29.51/0.0547
& 30.28/0.0546
& 29.11/0.0021
& \second{34.37}/0.1777
& \second{34.45}/0.1117
& 30.15/0.0937
& 30.26/0.0000 \\

& Average filter
& 30.68/0.2734
& 30.69/0.0456
& 31.34/0.0473
& 29.62/0.0135
& 28.91/0.2725
& 28.92/0.0344
& 29.52/0.0661
& 26.35/0.0346 \\

& JPEG
& {32.47}/0.2168
& 32.14/0.0820
& 32.54/0.1122
& 31.52/0.0030
& 31.72/0.2148
& 31.89/0.0547
& 29.33/0.1248
& 29.14/0.0020 \\
\hline

\multirow{5}{*}{\rotatebox[origin=c]{90}{Learning-based}}
& FAADW
& 31.02/0.2848
& 30.89/0.1503
& 31.78/0.3188
& 30.35/0.2946
& 28.75/0.3122
& 28.43/0.1389
& 26.21/0.3196
& 26.17/0.2720 \\

& HIWANet
& 31.76/0.2818
& 31.25/0.1753
& \second{32.58}/0.2688
& 31.62/0.2592
& 29.43/0.3048
& 30.97/0.2123
& 28.53/0.2725
& 30.28/0.2649 \\

& Diffusion Attack
& 31.72/0.2720
& \second{33.17}/\second{0.1909}
& 30.54/0.4039
& \second{34.64}/\second{0.4000}
& 30.83/0.2700
& 32.66/\second{0.2139}
& \second{30.36}/\second{0.4236}
& \second{34.02}/0.3210 \\

& UnMarker
& \second{32.64}/\second{0.2849}
& 32.41/0.1211
& 30.76/\second{0.4318}
& 30.05/0.3536
& 30.54/\second{0.3433}
& 30.36/0.1567
& 30.30/0.4200
& 30.34/\second{0.3948} \\

\cline{2-10}
& \textbf{Ours}
& \best{42.48/0.3135}
& \best{45.14/0.2363}
& \best{40.72/0.4496}
& \best{39.84/0.5333}
& \best{43.01/0.3740}
& \best{44.27/0.2578}
& \best{42.21/0.4388}
& \best{38.81/0.4667} \\
\hline

\end{tabular*}

\caption{Quantitative Comparison of different attack methods across various watermarking
methods on the CelebA and ImageNet datasets. \textbf{Bold} and \underline{underlined} values denote the best and second-best results, respectively.}
\label{tab:main_comparison}
\end{table*}
All experiments are implemented in PyTorch 2.5 and trained on a single NVIDIA GeForce RTX 4090 GPU. We optimize FDDWAN using Adam~\cite{kingma2014adam} with $\beta_1=0.9$, $\beta_2=0.999$, an initial learning rate of $1\times10^{-4}$, and a batch size of 2. The number of diffusion steps is set to $T = 1000$, and the model is trained for 500K optimization iterations. FDDWAN is evaluated on face images of CelebA dataset, as well as general natural images from ImageNet. All images are resized to 256×256. Furthermore, we evaluate four representative watermarking schemes: transform-domain DCT~\cite{sisaudia2022dct}, orthogonal moment-based PHFMs~\cite{phfm}, deep learning-based HiDDeN~\cite{zhu2018hidden} and StegaStamp~\cite{tancik2020stegastamp}. To comprehensively evaluate FDDWAN, we compare it with five conventional attacks—Gaussian noise (variance 0.001), salt-and-pepper noise (density 0.005), speckle noise (variance 0.005), average filtering with a 3×3 kernel, and JPEG compression with a quality factor of 50—as well as four representative learning-based methods: FAADW~\cite{chen2025faadw}, HIWANet~\cite{wang2024hiwanet}, Diffusion Attack~\cite{zhao2024vaeattack} and UnMarker~\cite{kassis2025unmarker}. For a fair comparison, all methods are evaluated using identical test images. We evaluate attack performance using PSNR~\cite{wang2003psnr},
SSIM~\cite{wang2004ssim}, and BER~\cite{ali2013ber}. Higher PSNR and SSIM
values indicate better visual fidelity, whereas, for binary watermark messages, a BER close to 0.5 indicates
random guessing and therefore stronger watermark removal.\begin{figure}[t]
\centering
\includegraphics[width=1.0\columnwidth]{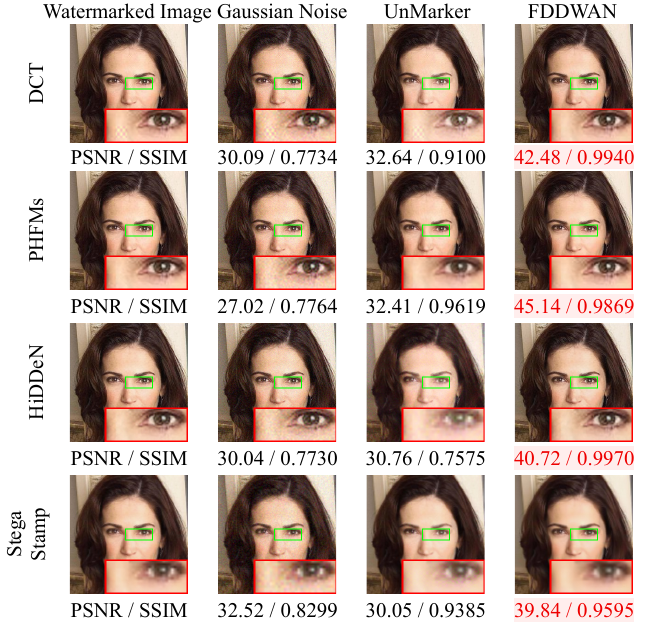} 
\caption{Qualitative comparison of Gaussian noise, UnMarker, and the
proposed FDDWAN.}
\label{fig:fig3}
\end{figure}
\subsection{Visual Quality and Attack Effectiveness}
As shown in Table~\ref{tab:main_comparison}, conventional attacks generally
provide limited watermark removal capability while causing noticeable
image degradation. Their PSNR values mostly fluctuate around $30$ dB,
indicating that fixed operations such as noise addition, filtering, and
JPEG compression inevitably disturb image content. Meanwhile, their BER
values remain relatively low in most cases. In particular, the BER values
against PHFMs and StegaStamp are often below $0.1$, and several attacks
produce values close to zero, showing that most embedded watermark bits
can still be correctly recovered. Although some conventional attacks
achieve moderately higher BER against DCT, their overall attack
performance remains inconsistent across different watermarking schemes.

Learning-based attacks generally achieve stronger watermark disruption,
particularly against HiDDeN and StegaStamp. UnMarker and Diffusion Attack,
for example, achieve BER values above $0.4$ against HiDDeN in several
settings. However, their effectiveness remains sensitive to the target
watermarking method, especially for PHFMs. Moreover, their PSNR values
mostly remain within $26$--$34$~dB, indicating that stronger watermark
disruption is often accompanied by noticeable visual degradation.

In contrast, FDDWAN achieves the highest PSNR and BER in all eight
dataset--watermarking combinations. It consistently increases BER across
both conventional and learning-based watermarking schemes while preserving
substantially higher image quality. In particular, FDDWAN improves the BER
over UnMarker in every evaluated setting and simultaneously increases
PSNR by a considerable margin. These results demonstrate that FDDWAN
effectively alleviates the common trade-off between watermark removal
strength and visual fidelity.

The qualitative results in Fig.~\ref{fig:fig3} further support the
quantitative comparison. Gaussian noise introduces visible grain-like
perturbations, whereas UnMarker causes noticeable smoothing and loss of
local details. In comparison, FDDWAN better preserves facial structures
and fine-grained textures, particularly around the enlarged eye regions.
Across the four visualized watermarking methods, FDDWAN consistently
achieves PSNR values of approximately $40$~dB and SSIM values above $0.95$.
Together with the consistently higher BER values, these results show that
FDDWAN effectively disrupts embedded watermarks while maintaining high
perceptual similarity to the corresponding watermarked images.
\subsection{The Generalizability of FDDWAN}
\begin{figure}[t]
\centering
\includegraphics[width=0.9\columnwidth]{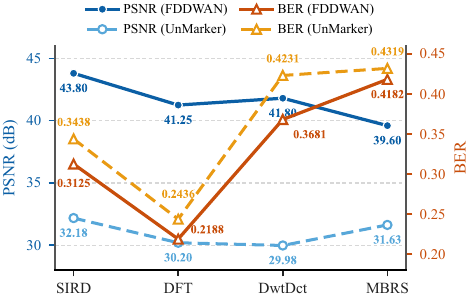} 
\caption{Comparison between UnMarker and our FDDWAN on the CelebA dataset for four unseen watermarking methods.}
\label{fig:gen}
\end{figure}

\paragraph{Generalization across post-hoc watermarking methods.}
We first evaluate FDDWAN on four post-hoc invisible watermarking
methods not included in the training set, including
SIRD~\cite{abraham2019imperceptible},
DFT~\cite{urvoy2014dft}, DWT-DCT~\cite{alhaj2007dwtdct}, and
MBRS~\cite{jia2021mbrs}. For each method, 500 images are randomly
selected from the CelebA test set for watermark embedding. As shown in
Fig.~\ref{fig:gen}, FDDWAN consistently achieves over 10~dB higher PSNR
than UnMarker while maintaining comparable BER across all four methods.
Although UnMarker attains higher BER, FDDWAN preserves better
image quality and achieves a more favorable trade-off between watermark disruption and visual fidelity. These results demonstrate the
generalization ability of FDDWAN across different conventional invisible
watermarking schemes.
\paragraph{Generalization across in-generation watermarking methods.}
To examine whether FDDWAN extends beyond post-hoc watermarking, we
further evaluate it on Stable
Signature~\cite{fernandez2023stable}, where the watermark is integrated
into the image generation process. To further assess perceptual similarity, we additionally employ
LPIPS~\cite{zhang2018lpips}, where a lower value indicates better visual
fidelity.
As shown in Table~\ref{tab:stable_signature}, FDDWAN achieves a BER of
0.42, close to the 0.44 obtained by Diffusion Attack and higher than the 0.31
of UnMarker. Meanwhile, it obtains the best PSNR and SSIM scores of
35.52~dB and 0.9257, respectively, together with a LPIPS score of
0.0344. 

Overall, FDDWAN maintains competitive watermark removal
performance while preserving high perceptual quality, showing a balanced
trade-off.
\begin{table}[t]
\centering
\small
\setlength{\tabcolsep}{5.5pt}
\renewcommand{\arraystretch}{1.15}

\begin{tabular}{lcccc}
\toprule
Method & BER & PSNR$\uparrow$ & SSIM$\uparrow$ & LPIPS$\downarrow$ \\
\midrule
UnMarker
& 0.31
& 25.54
& 0.7860
& \underline{0.0180} \\

Diffusion Attack
& \textbf{0.44}
& 30.01
& 0.8886
& 0.1178 \\

FDDWAN
& \underline{0.42}
& \textbf{35.52}
& \textbf{0.9257}
& \underline{0.0344} \\
\bottomrule
\end{tabular}
\caption{Quantitative comparison of different attacks on the
generative watermarking method Stable Signature.}
\label{tab:stable_signature}
\end{table}
\subsection{Ablation Studies}
\begin{table}[t]
\centering
\begin{tabular*}{0.82\columnwidth}{
    @{\extracolsep{\fill}}lcc@{}
}
\toprule
Method & PSNR$\uparrow$ & BER \\
\midrule
w/o DCB       & 33.14 & 0.0834 \\
w/o HFAB      & 29.14 & 0.1196 \\
w/o Low-query & 36.04 & 0.1208 \\
Ours          & \textbf{45.14} & \textbf{0.2363} \\
\bottomrule
\end{tabular*}
\caption{Ablation results for WFPAM on CelebA using the PHFMs watermarking scheme.}
\label{tab:wfpam_ablation}
\end{table}

\begin{table}[t]
\centering
\begin{tabular*}{0.82\columnwidth}{
    @{\extracolsep{\fill}}lcc@{}
}
\toprule
Configuration & PSNR$\uparrow$ & BER \\
\midrule
DM              & 30.86 & 0.0921 \\
RDM             & 32.37 & 0.1142 \\
RDM + D-L       & 40.85 & 0.1457 \\
RDM + D-H       & 42.58 & 0.1249 \\
RDM + D-L + D-H & \textbf{45.14} & \textbf{0.2363} \\
\bottomrule
\end{tabular*}
\caption{Ablation results for FRDAM on CelebA using the PHFMs watermarking scheme. DM and RDM denote the
vanilla diffusion model and the residual diffusion model, respectively.
D-L and D-H denote diffusion refinement of the low- and high-frequency
subbands, respectively.}
\label{tab:frdam_ablation}
\end{table}

\subsubsection{Ablation Study on WFPAM.}

We evaluate the contribution of each component in WFPAM on CelebA by
individually removing DCB, HFAB, and Low-query. The results are reported
in Table~\ref{tab:wfpam_ablation}. Removing any component decreases both
PSNR and BER, confirming that all three components contribute to the
overall performance. Removing HFAB produces the largest reduction in
PSNR, indicating that HFAB plays an important role in preserving textures
and local details. Removing DCB also leads to a substantial performance
decrease, demonstrating the importance of processing the low-frequency
subband for suppressing robust watermark signals. The performance
degradation caused by removing Low-query further indicates that guidance
from the low-frequency subband facilitates watermark disruption in the
high-frequency subbands. 

\subsubsection{Ablation Study on FRDAM. }

Table~\ref{tab:frdam_ablation} compares DM, RDM, and different
frequency-specific diffusion configurations on CelebA. Replacing DM with
RDM improves both PSNR and BER, indicating that residual prediction
provides a more suitable learning objective than direct image refinement.
Introducing D-L yields a larger improvement in BER, whereas D-H produces
a higher PSNR. This difference reflects the respective roles of diffusion
refinement in the low- and high-frequency subbands: the former primarily
enhances watermark suppression, whereas the latter contributes more to
visual detail preservation. Combining RDM, D-L, and D-H achieves the best overall
performance, demonstrating that the two frequency-specific refinement
branches are complementary.

\section{Conclusion}

In this paper, we propose FDDWAN, a frequency-domain diffusion framework
for invisible watermark removal. WFPAM first decouples the low- and
high-frequency wavelet subbands and suppresses watermark-related features
within these subbands. FRDAM then removes the remaining watermark residuals
through frequency-specific diffusion refinement. Experiments on CelebA and
ImageNet demonstrate that FDDWAN achieves a favorable balance between
watermark removal effectiveness and visual quality, while the ablation
results verify the contributions of both modules. However, FRDAM employs
two diffusion models to process the low- and high-frequency subbands
separately, and the iterative sampling procedures introduce substantial
computational overhead and slow inference. This limitation reduces the
practicality of FDDWAN in real-time and resource-constrained scenarios.
Future work will focus on reducing the number of sampling steps and the
computational cost through accelerated samplers, model distillation, and
a shared diffusion architecture.
\bibliography{aaai2027}
\end{document}